\documentclass[runningheads]{llncs}

\usepackage{eccv}

\usepackage{eccvabbrv}

\usepackage{graphicx}
\usepackage{booktabs}
\usepackage{multirow}
\usepackage[accsupp]{axessibility}  

\usepackage{hyperref}

\usepackage{orcidlink}

\begin{document}

\title{Learning Neural Deformation Representation for 4D Dynamic Shape Generation} 

\titlerunning{Neural Deformation Representation for 4D Shape Generation}

\author{Gyojin Han\orcidlink{0009-0002-7905-0045} \and
Jiwan Hur\orcidlink{0009-0003-7252-038X} \and
Jaehyun Choi\orcidlink{0000-0002-9183-761X} \and Junmo Kim\orcidlink{0000-0002-7174-7932}}

\authorrunning{G. Han et al.}

\institute{Korea Advanced Institute of Science and Technology\\
\email{\{hangj0820, jiwan.hur, chlwogus, junmo.kim\}@kaist.ac.kr}}

\maketitle

\begin{abstract}
Recent developments in 3D shape representation opened new possibilities for generating detailed 3D shapes. Despite these advances, there are few studies dealing with the generation of 4D dynamic shapes that have the form of 3D objects deforming over time. To bridge this gap, we focus on generating 4D dynamic shapes with an emphasis on both generation quality and efficiency in this paper. HyperDiffusion, a previous work on 4D generation, proposed a method of directly generating the weight parameters of 4D occupancy fields but suffered from low temporal consistency and slow rendering speed due to motion representation that is not separated from the shape representation of 4D occupancy fields. Therefore, we propose a new neural deformation representation and combine it with conditional neural signed distance fields to design a 4D representation architecture in which the motion latent space is disentangled from the shape latent space. The proposed deformation representation, which works by predicting skinning weights and rigid transformations for multiple parts, also has advantages over the deformation modules of existing 4D representations in understanding the structure of shapes. In addition, we design a training process of a diffusion model that utilizes the shape and motion features that are extracted by our 4D representation as data points. The results of unconditional generation, conditional generation, and motion retargeting experiments demonstrate that our method not only shows better performance than previous works in 4D dynamic shape generation but also has various potential applications. 

\end{abstract}

\section{Introduction}
\label{sec:intro}
In the fields of 3D computer vision and graphics, research on 3D shape representation has developed continuously along with significant interest. 
In particular, recently proposed innovative 3D representations such as neural occupancy fields \cite{Mescheder_2019_CVPR}, neural signed distance fields (SDFs) \cite{Park_2019_CVPR}, neural radiance fields (NeRFs) \cite{mildenhall2020nerf}, and Gaussian splitting \cite{kerbl20233d} enabled the expression and manipulation of highly detailed 3D shapes. 
As the development of 3D representations has progressed, research for the generation of 3D shapes has also accelerated and various 3D generation methods \cite{kleineberg2020adversarial, Chou_2023_ICCV, Erkoc_2023_ICCV} have been proposed. 

However, the environments of the real world are not composed of static 3D objects but rather of 3D objects that deform along the time dimension, essentially forming 4D objects. 
Despite the importance of handling 4D objects in various application areas such as virtual reality (VR), augmented reality (AR), and game development, 
studies for generating 4D objects have not been sufficiently explored compared to those on 3D. Therefore, we focus on the task of 4D dynamic shape generation in this paper. 
Previously, HyperDiffusion\cite{Erkoc_2023_ICCV} attempted to learn a diffusion model that directly generates optimized weights of 4D occupancy fields. 
To explain in more detail, HyperDiffusion requires the process of fitting one occupancy field for each 4D shape data sample.
After the fitting process for all data samples is completed, the diffusion model is learned using the weight parameters of the occupancy fields as a dataset, and denoising is performed in the weight space during sampling to directly generate optimized weight parameters.
HyperDiffusion achieved improved quality and finer details for 4D generation compared to a voxel-based diffusion baseline. 
However, when 4D shapes are generated in the form of 4D occupancy fields where the representation for motion is not separate from the shape representation, marching cubes mesh extraction \cite{10.1145/37401.37422} is required for every frame to convert occupancy fields to mesh sequences. This not only causes an extremely slow rendering speed but also makes it difficult to maintain temporal consistency because the vertices and faces of the triangle mesh of each frame are newly defined.

To address these issues, we propose a new neural deformation representation that effectively expresses deformations separated from shapes, while also being specialized for the 4D shape generation task. 
Our key idea is to make the deformation representation understand the whole structure of the shapes by predicting skinning weights \cite{kavan2014direct} corresponding to multiple parts of the shapes to perform transformations. 
Unlike previous 4D representation models focused on reconstruction tasks, which modeled the trajectory \cite{Niemeyer_2019_ICCV, Jiang_2021_CVPR} or direct movement \cite{Tang_2021_CVPR} of each 3D point separately for the representation of deformation, our deformation representation can use the information of the entire structure obtained from mesh sequences for 4D shape generation by expressing per-part transformations continuously across the time. 
In addition, we can encode 4D shapes to latent vectors by concatenating shape features and motion features following the extraction of them.
We use these latent vectors as data points to train a diffusion model. Then, we can complete the entire 4D shapes from the generated latent vector samples by using them as modulation vectors\cite{pmlr-v162-dupont22a, Chan_2021_CVPR, Mehta_2021_ICCV} for the representations at test time. 
Through unconditional generation experiments, we demonstrate that our method efficiently generates 4D shapes with significantly higher diversity, and temporal consistency compared to existing methods. Experiments on motion retargeting utilizing the disentangled shape latent space and motion latent space, or conditional generation experiments taking point clouds as a condition, demonstrate the scalability of our proposed representation and diffusion modeling.

The contributions of this work can be summarized as follows:
\begin{itemize}
    \item We propose a novel neural deformation representation that represents the deformation of 3D objects continuously across time. Our proposed representation works by predicting skinning weights and rigid transformations for each part, giving it an advantage over existing methods in understanding the structure of shapes.
    \item We design a diffusion model that utilizes latent vectors encoded from 4D dynamic shapes using the proposed representation. Our method ensures fast rendering speed and temporal consistency by separating the shape latent space and motion latent space to eliminate the need for the mesh to be redefined in every frame.
    \item We perform experiments on various tasks, not just unconditional 4D generation, but also conditional generation and motion retargeting, to demonstrate potential applications of our work.
\end{itemize}

\section{Related Work}
\label{sec:relwork}

\subsection{4D Representations}
Various 3D representations \cite{Mescheder_2019_CVPR, Park_2019_CVPR, mildenhall2020nerf, kerbl20233d} for expressing static 3D objects have achieved significant success, showing promising results in both quantitative and visual quality aspects. However, these methods do not consider the representation of deformations, making them unsuitable for representing dynamic 4D objects. To address this issue, various 4D representation methods have been proposed. First, a branch of work introduced morphable 3D parametric models designed to represent detailed 4D objects in domain-specific areas. Examples include A Skinned Multi-Person Linear Model (SMPL) \cite{10.1145/2816795.2818013} for representing 3D humans and MANO \cite{MANO:SIGGRAPHASIA:2017} for representing 3D hands. While these models have achieved remarkable success as domain-specific representations, they suffer from limited expressiveness due to their reliance on template meshes, and they can only represent a limited category of objects. Therefore, as a 4D representation method, morphable 3D parametric models \cite{10.1145/2816795.2818013, MANO:SIGGRAPHASIA:2017, li2017learning} and their variants \cite{Ma_2020_CVPR, Jiang_2022_CVPR} do not align well with our method which aims for 4D generation of general objects. 
Model-free 4D representation methods \cite{Niemeyer_2019_ICCV, Tang_2021_CVPR, Lei_2022_CVPR}, that do not rely on manually created template meshes have also been proposed. They can represent general 4D objects by modeling the deformations of 3D objects between frames.
Occupancy Flow (OFlow) \cite{Niemeyer_2019_ICCV} adopts Neural ODE \cite{NEURIPS2018_69386f6b} for the architecture of the velocity field which outputs the velocity of each point in 3D coordinates at a given time. LPDC \cite{Tang_2021_CVPR} uses a Multi-layer Perceptron (MLP) to represent correspondences in parallel. However, these methods do not have sufficient interaction between the mesh representations and the motion representations. Therefore, when they are used for 4D shape generation, they can generate motions that are far from the shapes. In addition, since they model the trajectory or movement of each separate point, the correlation between the movements of points belonging to the same part may be insufficient. 
Therefore, we propose a 4D representation specialized for 4D generation tasks by predicting skinning weights of vertex points for the parts of the object and generating corresponding transformations.

\subsection{Generative Models for 3D and 4D Shapes}

Unlike early 3D generation studies\cite{pmlr-v80-achlioptas18a, valsesia2018learning, Shu_2019_ICCV, Yang_2019_ICCV} that focused on point cloud generation, recent studies aim to generate continuous implicit neural fields using generative adversarial networks (GANs)\cite{NIPS2014_5ca3e9b1} or diffusion models\cite{NEURIPS2020_4c5bcfec}, as research on 3D representations continues. Following this trend, numerous studies on 3D generation \cite{kleineberg2020adversarial, Chou_2023_ICCV, Erkoc_2023_ICCV, hui2022neural, NEURIPS2022_cebbd24f} have been conducted.
Despite the significant interest in 3D shape generation, research on generating 4D objects, which consists of a sequence of 3D objects has not been actively explored. Recently, HyperDiffusion\cite{Erkoc_2023_ICCV} attempted to learn a diffusion model that generates optimized weights of 4D occupancy fields directly. 
However, HyperDiffusion suffers from slow rendering speed and poor temporal consistency by using 4D occupancy fields where the representations of shape and motion are not separated.
Therefore, we present a deformation representation separated from the shape representation and show the potential for various applications such as motion retargeting.

\section{Method}
\label{sec:method}
In this section, we introduce our method for 4D dynamic shape generation. First, we describe a neural representation for 4D shapes, which constructs a latent space for 4D shapes, allowing each deforming object to be represented in the form of a latent vector. Next, we introduce the training process of a diffusion model that utilizes the extracted latent vectors.

\begin{figure}[t]
  \centering
\includegraphics[width=1.0\linewidth]{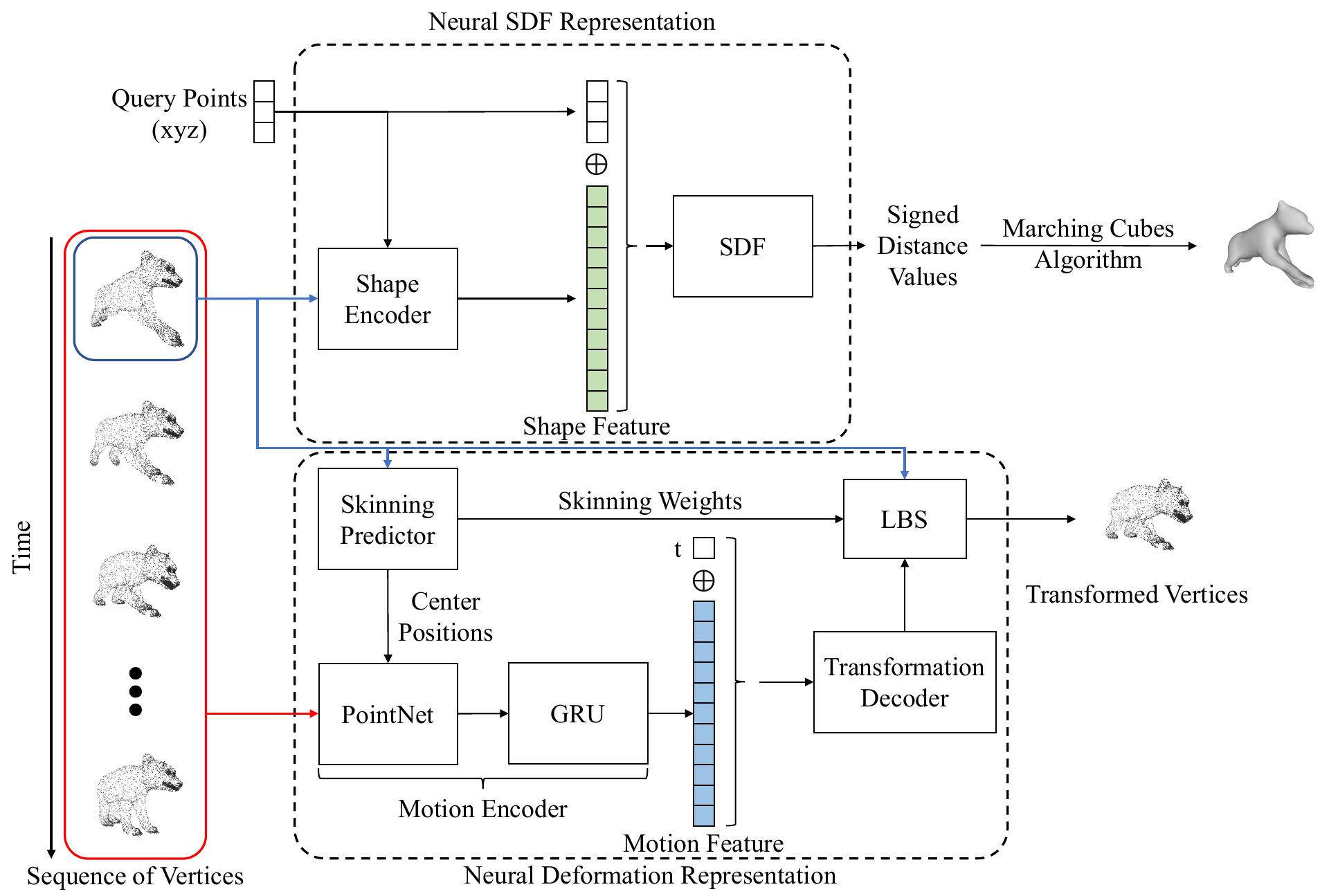}
  \caption{An overview of our proposed 4D dynamic shape representation. The neural SDF representation is trained to minimize the distance between the predicted SDF values and the ground truth values, while the neural deformation representation is trained to minimize the distance between the predicted vertex sequence positions and the ground truth positions.
  }
  \label{method_representation}
\end{figure}

\subsection{Neural Representation for 4D Shapes}
To address the issues arising from redefining meshes for every frame in the previous work, we propose a model that generates per-vertex animation.
Per-vertex animation defines mesh sequences by maintaining the information of the vertices defined in the first frame and the faces connecting these vertices while representing subsequent frames through the offset of vertex positions. 
Therefore, we introduce a new neural representation architecture consisting of two parts: an SDF representation to encode mesh surfaces and a deformation representation to encode the deformations of the mesh vertices. The overall architecture of the proposed representation is described in \cref{method_representation}.

\noindent \textbf{Neural SDF representation.}
In this work, we use neural signed distance fields (SDFs) as the 3D representation to continuously parameterize mesh surfaces. 
However, not only directly generating SDFs using generative models such as diffusion models is challenging, but it also requires the SDF fitting process for all training data, which is incredibly time-consuming.
Therefore, we adopt a strategy of training a conditional SDF representation and extracting a shape feature for each mesh to utilize it as a modulation vector. 
Following the previous works \cite{peng2020convolutional}, we extract shape features of meshes through a shape encoder $f_{p}$ with a variant of PointNet \cite{Qi_2017_CVPR} architecture and then concatenate these with 3D query points to input them into an SDF model $f_{SDF}$ which outputs signed distance values. 
Our shape encoder includes an additional auto-encoder, which is aimed at encoding the tri-plane shape feature as a simpler 1-dimensional vector.
Then, by utilizing the shape feature of each mesh as a modulation for the SDF representation, we can represent the meshes as 1-dimensional vectors.
The SDF representation can be trained to minimize the $l_1$ distance loss between predicted signed distance values $\hat{\mathbf{d}}$ and ground truth signed distance values $\mathbf{d}\in \mathbb{R}^{P}$ for a total of P query points $\mathbf{x}\in \mathbb{R}^{P\times 3}$. The loss, $\mathcal{L}_{SDF}$, is calculated as below:
\begin{equation}
\begin{split}
\mathcal{L}_{SDF} &= \frac{1}{|\mathcal{B}|}\sum_{(\textbf{V}^{0},( \mathbf{x},\mathbf{d}))\in \mathcal{B}}  \left\| \hat{\mathbf{d}} - \mathbf{d}\right\|_1 \\
s.t. \quad \hat{\mathbf{d}} &= f_{SDF}(\text{concat}(\textbf{x}, f_{p}(\mathbf{V}^0, \mathbf{x})),
\end{split}
\end{equation}
where $\mathbf{V}^0\in\mathbb{R}^{N\times{3}}$ is a total of $N$ vertex positions of the initial mesh, $\mathbf{V} = \{\mathbf{V}^0, ..., \mathbf{V}^{T-1}\}$ is the entire sequence of vertex positions for a 4D shape of length $T$, and $\mathcal{B}$ denotes a batch for each training step.

\noindent \textbf{Neural deformation representation.}
Similar to the case of SDFs described above, we propose a neural deformation representation conditioned on a series of vertex positions to represent deformations as motion features.
Our proposed neural deformation representation begins with the idea of making a neural representation emulate the process of generating skeletal animations, which is a method often used by human animators to create animations.
In the generation process of skeletal animation using linear blend skinning (LBS) \cite{kavan2014direct}, each frame's vertex positions are calculated using rigid transformations corresponding to each bone. 
By applying LBS, each vertex moves according to the bones' rigid transformations and then is scaled based on the skinning weights corresponding to each bone.

Following this idea, we design an architecture that works in a different way from existing 4D representations \cite{Niemeyer_2019_ICCV,Tang_2021_CVPR, Lei_2022_CVPR}. We make our neural deformation representation predict skinning weights and rigid transformations from vertex sequences and map from the deformations to modulation vectors.
Our neural deformation representation $g$ is composed of three modules, 1) a skinning predictor, 2) a motion encoder, and 3) a transformation decoder. 

The skinning predictor $g_{skin}$ has an architecture similar to the shape encoder of previously introduced SDF representation, but it includes an additional dense classifier that performs point segmentation. It predicts the weights $w \in[0,1]^k$ of each vertex point for $k$ parts from the initial mesh vertices of animation, where $k$ is the total number of the parts. After the prediction of the skinning weights, the center of the part $i$, $\mathbf{C}_i$, can be calculated as below:

\begin{equation}
\mathbf{C}_i = \frac {\mathbf{W}_{i}\mathbf{V}^0}{\sum _{n=1}^{N} \mathbf{W}_{i,n}} \qquad s.t. \quad \mathbf{W} = g_{skin}(\mathbf{V}^0),
\end{equation}
where $N$ is the total number of the vertices, $\mathbf{W}\in[0,1]^{k\times{N}}$ is the weight matrix for all vertices, $\mathbf{W}_i \in[0,1]^{1\times{N}}$ is $i$-th row of matrix $\mathbf{W}$, and $\mathbf{W}_{i,n}$ denotes the $n$-th vertex's weight for part $i$. 
The motion encoder $g_{m}$ takes the entire sequence of vertex points as input and extracts motion features. It consists of a variant of PointNet encoder and a gated recurrent unit (GRU) \cite{cho2014learning} module. The encoder extracts tri-plane features for each $t$-th frame and then samples the point features at the center positions $\mathbf{C}  \in \mathbb{R}^{ k \times 3} $ of $k$ parts where $\mathbf{C}_i$ is $i$-th row of matrix $\mathbf{C}$. After that, the GRU processes the sequentially calculated features of $k$ parts to generate a motion feature that contains information about the entire motion.
The transformation decoder $g_{trans}$ has a simple Multi-Layer Perceptron (MLP) architecture and takes the concatenated value of the time $t$ and motion features as input to decode the transformation matrix at each time $t$. 
In other words, the transformation decoder serves as a neural representation for deformations that use motion features as modulation vectors.

Through the above process, we can get the center positions of $k$ parts, skinning weights of vertex points, and rigid transformations. Therefore, we can transform each vertex point continuously across time using LBS as in the equation below:
\begin{equation}
\begin{split}
\hat{\mathbf{V}}_{n}^{t} &= \sum _{i=1}^{k} \mathbf{W}_{i, n}(\mathbf{T}^{t}_{i}(\mathbf{V}_{n}^{0} - \mathbf{C}_{i})) \\ s.t. \quad \mathbf{T}^t &= g_{trans}(\text{concat}(t, g_{m}(\mathbf{V}, \mathbf{C}))),
\end{split}
\end{equation}
where $\hat{\mathbf{V}}^{t}$ denotes the predicted vertices at time $t$, $\mathbf{T}^t = \{\mathbf{T}_1^t, ..., \mathbf{T}_k^t\}$ is the set of rigid transformations inferred by the transformation decoder for all $k$ parts at time $t$, and $\mathbf{V}_n^t$ means the $n$-th vertex of $\mathbf{V}^t$.
Our deformation representation is trained to minimize correspondence loss. The correspondence information is obtained from the ground truth sequence of vertex positions. The correspondence loss, $\mathcal{L}_{corr}$, is calculated as follows:
\begin{equation}
\mathcal{L}_{corr} = \frac{1}{|\mathcal{B}|}\sum_{(\mathbf{V}, t)\in \mathcal{B}}  \left\| \hat{\mathbf{V}}^t - \mathbf{V}^t\right\|_l
\end{equation}
where $\mathcal{B}$ denotes a batch for each training step, and $l$ is the order of the norm.

\begin{figure}[t]
  \centering
\includegraphics[width=1.0\linewidth]{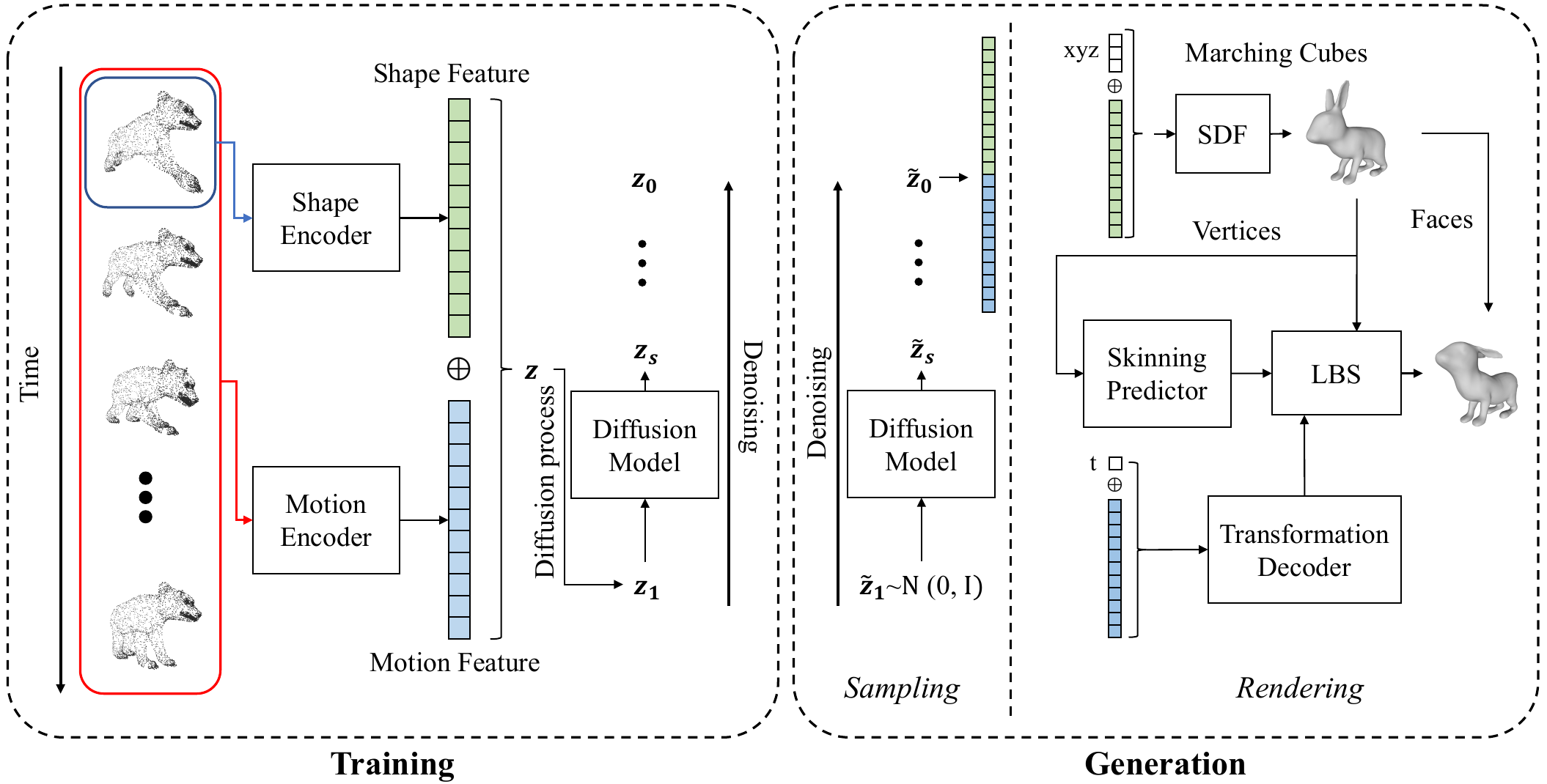}
  \caption{An overview of the training and generation process of the proposed diffusion model. We extract shape features and motion features from 4D shapes using the proposed neural representation, and then utilize concatenated latent vectors derived from them for the training of the diffusion model. During the generation process, the trained diffusion model generates latent vectors from Gaussian noise through the denoising process. These latent vectors are then split to obtain shape features and motion features, which are used as modulation vectors for decoding of the first frame's mesh and deformed vertices at each time.
  }
  \label{method_diffusion}
\end{figure}

\subsection{Diffusion Model for 4D Generation}
After the training of our proposed neural representation for 4D shapes is complete, we can encode all 4D shapes of the training dataset into shape features $\mathbf{s}$ and motion features $\mathbf{m}$ using the shape encoder $f_p$ of the SDF representation and the motion encoder $g_{m}$ of the deformation representation. We use 1-dimensional vectors $\mathbf{z}$, concatenated from $\mathbf{s}$ and $\mathbf{m}$, for training a diffusion model.

For the generation of novel 4D shapes, we utilize the diffusion model $\mathbf{\Omega}$ to approximate the ground truth latent vector distribution $p(\mathbf{z})$ and sample the latent vector $\Tilde{z}$. 
Specifically, for a continuous diffusion timestep $s \in [0,1]$, $z_0 \sim p(\mathbf{z})$ and $z_1 = \epsilon \sim \mathcal{N}(0,\mathbf{I})$ denote the sampled latent vector and Gaussian noise, respectively. Following the previous works~\cite{ramesh2022hierarchical,Chou_2023_ICCV}, the diffusion model $\mathbf{\Omega}$ learns to predict input data $z_0$ from the noise perturbed data $z_s = \sqrt{\Bar{\alpha}_s}z_0 + \sqrt{1-\Bar{\alpha}_s}\epsilon$ using the score matching loss function:
\begin{equation}
\mathcal{L}_{diff}=\left\|\mathbf{\Omega}(z_s,s)-z_0\right\|_2,
\label{eq:diffusion}
\end{equation}
where $\Bar{\alpha}_s \in [0,1]$ is a noise scheduling function which monotonically decreases with timestep $s$.
We use the DALLE-2~\cite{ramesh2022hierarchical} architecture for $\mathbf{\Omega}$ to deal with the 1-dimensional diffusion process.
After the training of $\mathbf{\Omega}$, we can sample a new latent vector $\Tilde{z}$ using iterative denoising processes such as DDPM~\cite{ho2020denoising} or DDIM~\cite{song2020denoising}.

Then, we split the generated latent vector $\Tilde{z}$ into a shape feature $\Tilde{s}$ and a motion feature $\Tilde{m}$.
We can use the generated shape feature $\Tilde{s}$ with the SDF model $f_{SDF}$ for marching cubes mesh extraction to generate mesh vertices $\Tilde{\mathbf{V}}^0$ for the first frame and faces $\Tilde{\mathbf{F}}$ of a new 4D shape. In addition, we can complete the 4D shape $(\Tilde{\mathbf{V}}, \Tilde{\mathbf{F}})$ by inferring the vertex positions for each time $t$ using the motion feature $\Tilde{m}$ as follows:
\begin{equation}
\begin{split}
\Tilde{\mathbf{V}}_{n}^{t} &= \sum _{i=1}^{k} \Tilde{\mathbf{W}}_{i, n}(\Tilde{\mathbf{T}}^{t}_{i}(\Tilde{\mathbf{V}}_{n}^{0} - \Tilde{\mathbf{C}}_{i})) \\ s.t. \quad \Tilde{\mathbf{T}}^t &= g_{trans}(\text{concat}(t, \Tilde{m})),
\end{split}
\end{equation}
where $\Tilde{\mathbf{W}}$ and $\Tilde{\mathbf{C}}$ denotes the skinning weights and center positions calculated using $\Tilde{\mathbf{V}}^0$, respectively.
We describe the training and generation process of the diffusion model in \cref{method_diffusion}.

\section{Experiments}
\label{sec:experiments}

In this section, we report the results of the various experiments conducted to demonstrate the effectiveness of our method. 
First, we evaluate the generation quality of our method by comparing both quantitative and qualitative results of the unconditional 4D shape generation with a baseline. We also demonstrate our method's advantages by comparing temporal consistency and rendering speed.
Next, we investigate the scalability of our method through experiments on conditional generation and motion retargeting. 
Finally, although we focus on the 4D generation task, we conduct a 4D point cloud completion experiment to demonstrate the representation power of the proposed deformation representation.

\subsection{Experimental Setting}
\noindent \textbf{Datasets.}
We use the DeformingThings4D\cite{Li_2021_ICCV} dataset for 4D shape generation and motion retargeting experiments. 
DeformingThings4D contains 1,772 synthetic animation sequences of animals. 
We evenly sample 16 frames across each entire animation sequence. Then, we sample 4,096 vertex positions for each frame of the animation sequences and use them to train our representation. 
Following previous work\cite{Erkoc_2023_ICCV}, we divide the dataset into 80\% for the training set, 5\% for the validation set, and 15\% for the test set.
For the conditional 4D shape generation, we sample 2,048 points from a mesh surface of each sequence's first frame and use this as the condition during the proposed model's training and testing phases.
For the 4D point cloud completion experiment, we use the D-FAUST\cite{Bogo_2017_CVPR} dataset which contains real 4D human scans. Each sequence is subsampled into point cloud trajectories of 17 frames across time and each frame consists of 300 points.

\noindent \textbf{Baselines.} 
To evaluate the generation quality of our method, we compare the unconditional generation results from our method with those of HyperDiffusion\cite{Erkoc_2023_ICCV}, a recently proposed unconditional 4D shape generation method. 
HyperDiffusion generates the weights of 4D neural occupancy fields directly through a diffusion model. We reproduced the unconditional generation results of HyperDiffusion using the official codebase provided by the authors.
For the 4D point cloud completion experiment,
we compare our method with the model-free dynamic 4D representation methods\cite{Fan_2017_CVPR, Mescheder_2019_CVPR, Niemeyer_2019_ICCV, Jiang_2021_CVPR, Tang_2021_CVPR, Lei_2022_CVPR}.

\noindent \textbf{Implementation details.} 
For the training of both the SDF representation and deformation representation, we use the Adam optimizer with a batch size of 4 and an initial learning rate of $1\times10^{-4}$, which is reduced by 50\% for every 125 epochs. When learning our deformation representation, we set $k$, a hyperparameter indicating the number of parts, to 40.
For the training of the diffusion model, we use the Adam optimizer with a batch size of 16 and a learning rate of $1\times10^{-5}$.
We train both representations for 2,000 epochs and the diffusion model for 4,000 epochs.
More details can be found in the supplementary material.

\begin{table}[t]
\renewcommand{\arraystretch}{1.0}
\renewcommand{\tabcolsep}{2.0mm}
\caption{4D shape generation quantitative results. The mesh extraction time is measured as the average execution time of the marching cubes algorithm using a single A100 GPU. We used 200 samples for time measurement.}
\centering
\begin{tabular}{c|cccc}
  \toprule
 Method & MMD$\downarrow$ & COV (\%)$\uparrow$ & 1-NNA (\%)$\downarrow$ & Time (s)$\downarrow$\\ 
  \midrule
 HyperDiffusion & 19.6 & 41.8 & 64.1 & 40.4\\
 Ours (Unconditional) & 16.4 & 52.8 & \textbf{61.9} & \textbf{8.1} \\ 
 Ours (Conditional) & \textbf{15.6} & \textbf{53.2} & 62.5 & 8.2 \\
   \bottomrule
\end{tabular}
\label{result-table-generation}
\end{table} 

\noindent \textbf{Evaluation metrics.}
Following previous works\cite{Yang_2019_ICCV, zeng2022lion, Luo_2021_CVPR, Chou_2023_ICCV, Erkoc_2023_ICCV}, we use three evaluation metrics, 1) Minimum Matching Distance (MMD), 2) Coverage (COV), and 3) 1-Nearest-Neighbor Accuracy (1-NNA), for 4D shape generation experiments. 
We measure the quality of the generated samples using MMD and their diversity with COV.
Additionally, 1-NNA (1-Nearest Neighbor Accuracy) is used to measure the similarity between the distributions of real samples and generated samples. For a detailed description of these metrics, please refer to the work of Yang \etal\cite{Yang_2019_ICCV}.
For all metrics, the distance between each pair of samples was calculated using the Chamfer Distance (CD) of 2,048 points sampled from the mesh surfaces. The same number of generated samples as real samples were used for evaluation and each sample is standardized with its mean and standard deviation. 
The assessment of the quality of generated samples is largely subjective and it is difficult to evaluate the quality using quantitative metrics alone. Therefore, we additionally provide qualitative results for the experiments.
For the 4D point cloud completion experiment, we use Intersection over Union (IoU), CD, and correspondence $l_2$-distance error (Corr) as evaluation metrics.

\subsection{Unconditional 4D Dynamic Shape Generation}
\begin{figure}[t]
  \centering
\includegraphics[width=1.0\linewidth]{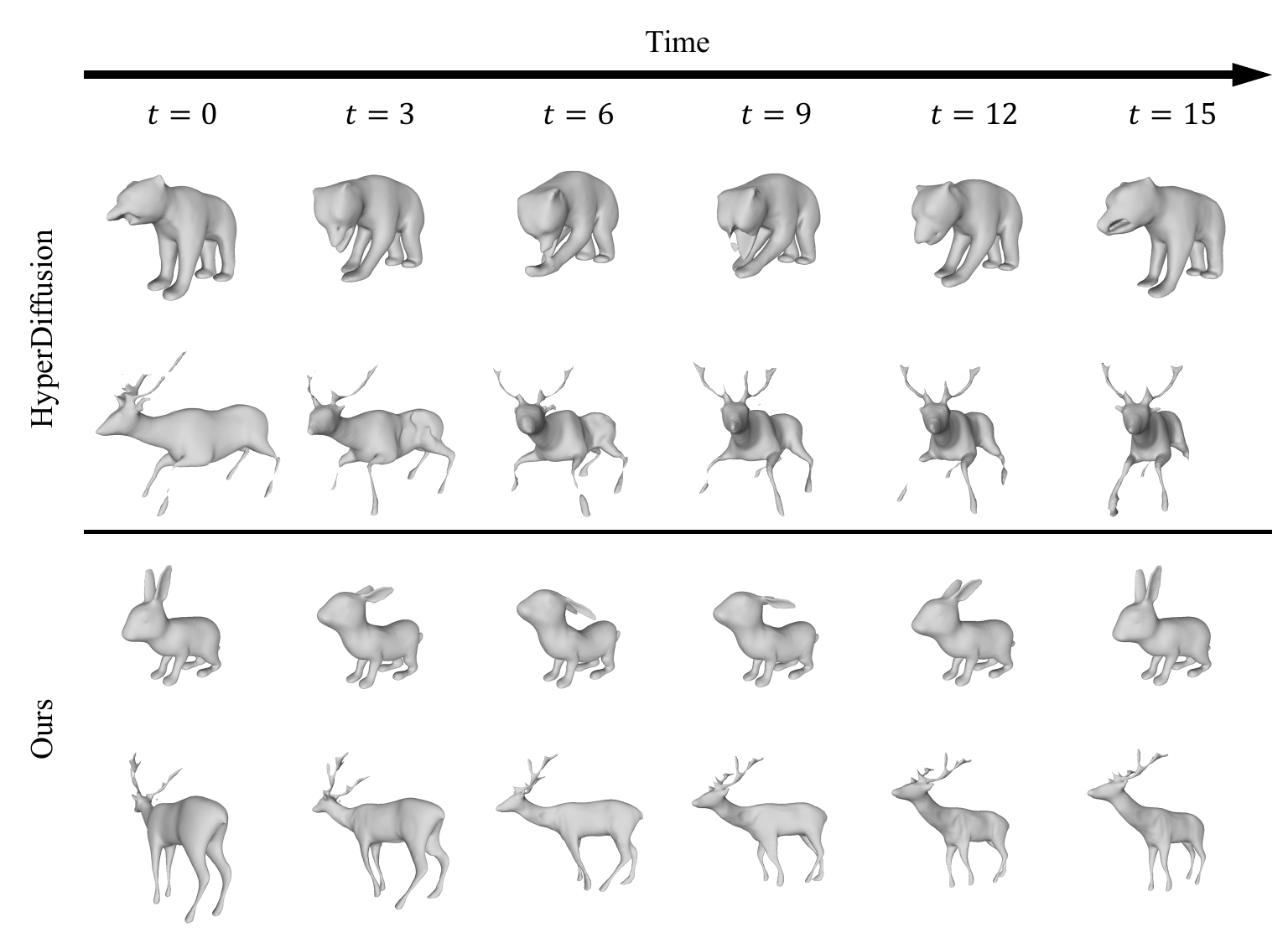}
  \caption{Unconditional 4D shape generation qualitative results. We visualize a total of 6 frames with a time interval of 3 frames from the generated 4D dynamic shapes.
  }
  \label{unconditional_generation}
\end{figure}

\begin{figure}[t]
  \centering
\includegraphics[width=0.6\linewidth]{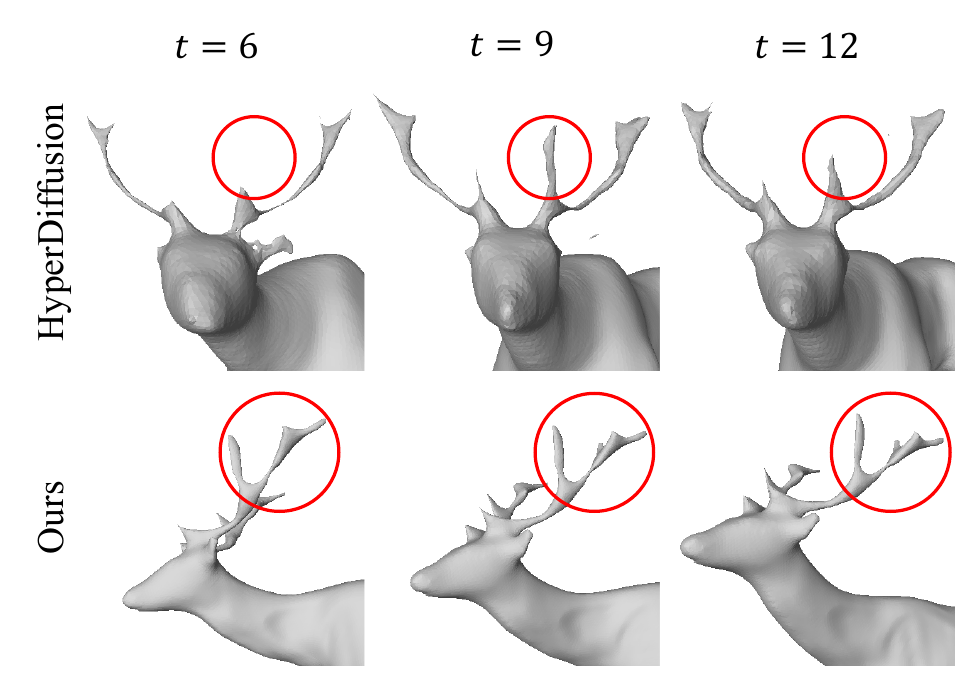}
  \caption{Temporal consistency comparison. In the case of HyperDiffusion, the generated sample suffers from a lack of temporal consistency, as parts of horns appear and disappear due to vertices and faces being redefined in every frame. However, the 4D shape generated by our method maintains excellent temporal consistency, even as the mesh is transformed by motion.
  }
  \label{temporal_consistency}
\end{figure}

We unconditionally generate animal animation sequences consisting of 16 frames using the proposed model trained on the DeformingThings4D dataset.
As shown in \Cref{result-table-generation}, our method achieves enhanced performance compared to the baseline in quantitative metrics.
In particular, the significant improvement in coverage compared to the baseline demonstrates that our method generates much more diverse shapes and motions.
In addition, we visualize the generated samples in \cref{unconditional_generation} to compare the visual quality, which cannot be fully captured by quantitative metrics.
The samples generated by our method not only have superior visual quality in each frame but also show significantly improved temporal consistency. 
Unlike HyperDiffusion, which requires mesh extraction at every time step, leading to vertices and faces being redefined in each frame, our method generates per-vertex animation defined by the mesh in the first frame and offsets of vertices.
As seen in \cref{temporal_consistency}, our method can generate samples that maintain temporal consistency even in highly detailed areas such as horns. 
Additionally, compared to HyperDiffusion, the proposed method ensures a much faster sampling time. 
Our method only needs to extract a mesh once for the first frame, and then deform the vertices from the mesh to generate meshes for the following frames.
In \Cref{result-table-generation}, we also present the experimental results for rendering time.

\subsection{Conditional 4D Dynamic Shape Generation}

\begin{figure}[t]
  \centering
\includegraphics[width=1.0\linewidth]{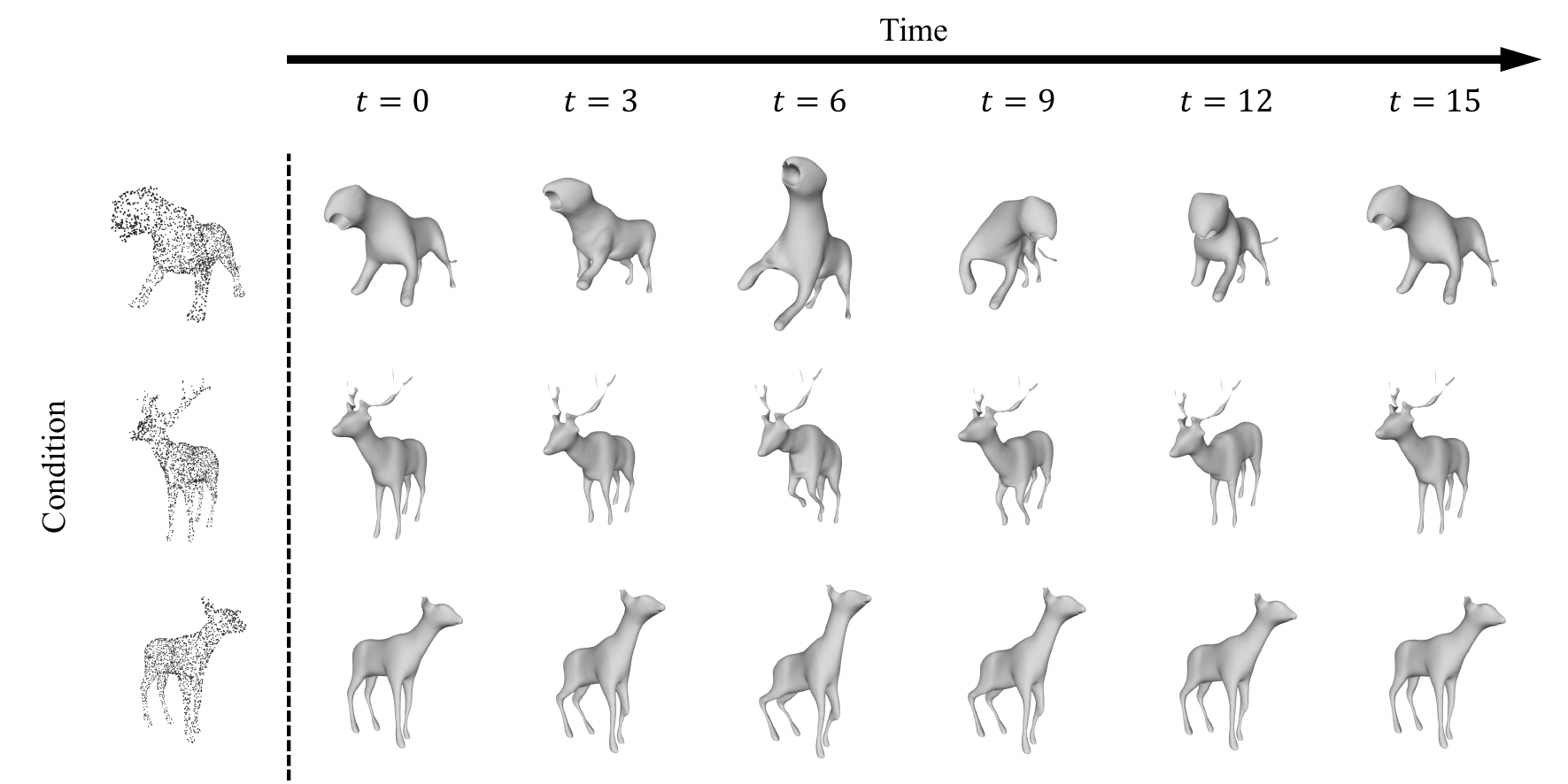}
  \caption{Conditional 4d generation qualitative results. We visualize a total of 6 frames with a time interval of 3 frames from the generated 4D dynamic shapes. In addition, we visualize the point clouds provided for the conditional generation of shape sequences.}
  \label{conditional_generation}
\end{figure}

To prove the scalability of our method, we conduct the first experiments on conditional 4D shape generation that have not been explored in previous work. 
We follow the mechanism proposed by the latent diffusion model\cite{Rombach_2022_CVPR}, utilizing an encoder for conditions and cross-attention layers for conditioning.
We attempt to specify the shape of the generated samples by providing point clouds as the condition. 
At the training and test stage, the point clouds are extracted from the first frame of each animation sequence.
In this setting, our model has to construct complete shapes from the provided point clouds and generate motions that match the shapes well. 
We report the quantitative results of conditional 4D shape generation experiments in \Cref{result-table-generation} with other experiments. 
Through quantitative results, we can confirm that our model can generate 4D shapes of similar quality with provided point clouds to those generated by unconditional generation.
The visualization results of conditional generation experiments in \cref{conditional_generation} show that completed shapes and generated motions match the input conditions well. 
Additionally, we believe that our method could be applied to various conditional generation tasks if datasets that include text labeling describing shapes or motions, or datasets of 4D shapes paired with images or videos, are collected.

\subsection{Motion Retargeting}

\begin{figure}[t]
  \centering
\includegraphics[width=1.0\linewidth]{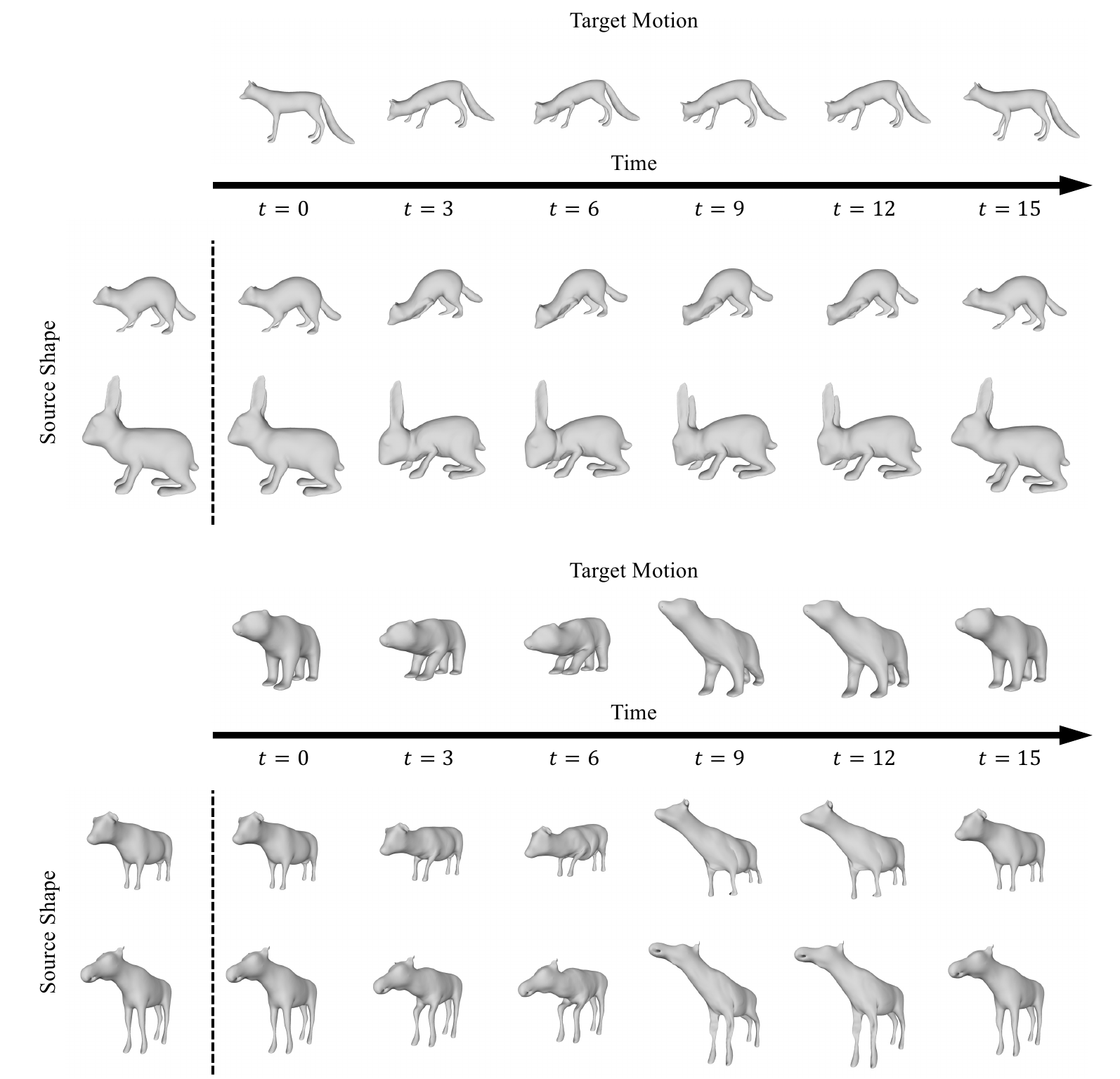}
  \caption{Motion retargeting results. We visualize motion-retargeted dynamic shapes by transferring target motions to two different source shapes each.
  }
  \label{motion_retargeting}
\end{figure}

Motion retargeting is a task that aims to transfer motions from one subject to another subject. 
Our model constructs a latent space where motion features and shape features are disentangled, allowing for easy motion retargeting.
Motion-retargeted shape sequences are achieved by generating the mesh of the first frame using the source shape features and then transforming this mesh using the target motion features.
All the 4D shapes and features used in this experiment are generated in the same settings as the unconditional 4D dynamic shape generation experiment.
We show the motion retargeting results in \cref{motion_retargeting}.
Even when there are significant differences in the source shapes from the original shape at the shape of each part (e.g., leg length, tail shape), we can confirm that the target motions are transferred well. Unlike the deformation modules of existing 4D representations, which rarely utilize information about shapes, our neural deformation representation predicts the skinning weights and center positions of all parts for the generated shapes. Therefore, our method enables deformations that are appropriate for each shape, even when utilizing shape features and motion features extracted from different samples.

\subsection{4D Point Cloud Completion} 

Although this paper focuses on the generation task rather than the reconstruction task, we conduct a 4D point cloud completion experiment to compare the representation power of our proposed deformation representation with the deformation modules of existing 4D representations. In \Cref{result-table-completion}
, we compare the results of 4D point cloud completion against existing 4D representation methods when replacing the deformation module of the state-of-the-art 4D representation method, CaDeX\cite{Lei_2022_CVPR}, with our deformation representation. The results demonstrate that although our proposed deformation representation targets 4D generation, the representation power of our method surpasses or is comparable to the deformation modules of existing 4D representations.

\begin{table}[t]
\renewcommand{\arraystretch}{1.0}
\renewcommand{\tabcolsep}{2.0mm}
\caption{4D point cloud completion results on D-FAUST dataset. The best and second-best scores are written in bold and underlined, respectively.}
\centering
\begin{tabular}{cc|ccc}
  \toprule
 \multicolumn{2}{c|}{Method} & IoU (\%)$\uparrow$ & CD$\downarrow$ & Corr$\downarrow$  \\
 \midrule
 \multirow{6}{*}{W/o. Corr. Supervision} &  PSGN-4D & - & 0.127 & 3.041 \\
 & ONet-4D & 66.6 & 0.140 & - \\
 & O-Flow  & 69.6 & 0.095 & 0.149 \\ 
 & LCR & 68.2 & 0.100 & - \\
 & LCR-F  & 69.9 & 0.094 & - \\
 & CaDeX  & 75.4 & 0.074 & 0.126 \\
 \midrule 
 \multirow{6}{*}{With Corr. Supervision} &  PSGN-4D & - & 0.119 & 0.131 \\
 & O-Flow & 72.3 & 0.084 & 0.117 \\
 & LPDC & 76.2 & 0.071 & 0.098 \\
 & CaDeX (NICE) & 75.6 & 0.070 & 0.104 \\
 & CaDeX (NVP) & \textbf{78.1} & \textbf{0.063} & \underline{0.095} \\
 & CaDeX (\textbf{Ours}) & \textbf{78.1} & \underline{0.064} & \textbf{0.093} \\
   \bottomrule
\end{tabular}
\label{result-table-completion}
\end{table}

\section{Limitations and Future Work}

In our method, we use LBS to deform the vertices with generated rigid transformations. While LBS facilitates efficient deformation of 3D models, it introduces specific issues, such as the candy wrapper effect, which unrealistically distorts the mesh, and its inability to capture non-rigid deformations perfectly. These limitations highlight the need for exploring more advanced skinning techniques, such as dual quaternion skinning\cite{10.1145/1409625.1409627} for our method, which can offer more accurate and realistic deformations, representing a promising direction for future work.

Although the DeformingThings4D dataset has the advantage of covering a very wide range of shapes, the number of shapes in the dataset is 59, which may be insufficient to learn a well-structured latent space. Therefore, if a large 4D object dataset containing more shapes and motions is presented, it will be possible to learn a model with a latent space that well represents diverse 4D dynamic shapes. 

\section{Conclusion}

In this paper, we propose a neural representation for 4D dynamic shapes and introduce a new approach to 4D shape generation utilizing a diffusion model with the proposed neural representation. 
Our approach guarantees quick rendering times and stable temporal consistency by dividing the shape latent space from the motion latent space, which removes the necessity to redefine the mesh in each frame.
In addition, the proposed deformation representation is specialized for 4D generation as it has a better understanding of the movement of the entire structure than existing 4D representations as it expresses per-part transformations continuously over time.
The unconditional 4D generation experiment demonstrates that our method is superior to existing methods in both qualitative and quantitative evaluations. Additionally, conditional generation and motion retargeting experiments demonstrate the potential of our method for various applications.

\subsubsection*{Acknowledgments.} This research was supported by a grant of the Korea Health Technology R\&D Project through the Korea Health Industry Development Institute (KHIDI), funded by the Ministry of Health \& Welfare, Republic of Korea (grant number : HI22C1496).

%
%
\bibliographystyle{splncs04}
\bibliography{main}
\clearpage
\appendix
\begin{figure}[h!]
  \centering
\includegraphics[width=0.6\linewidth]{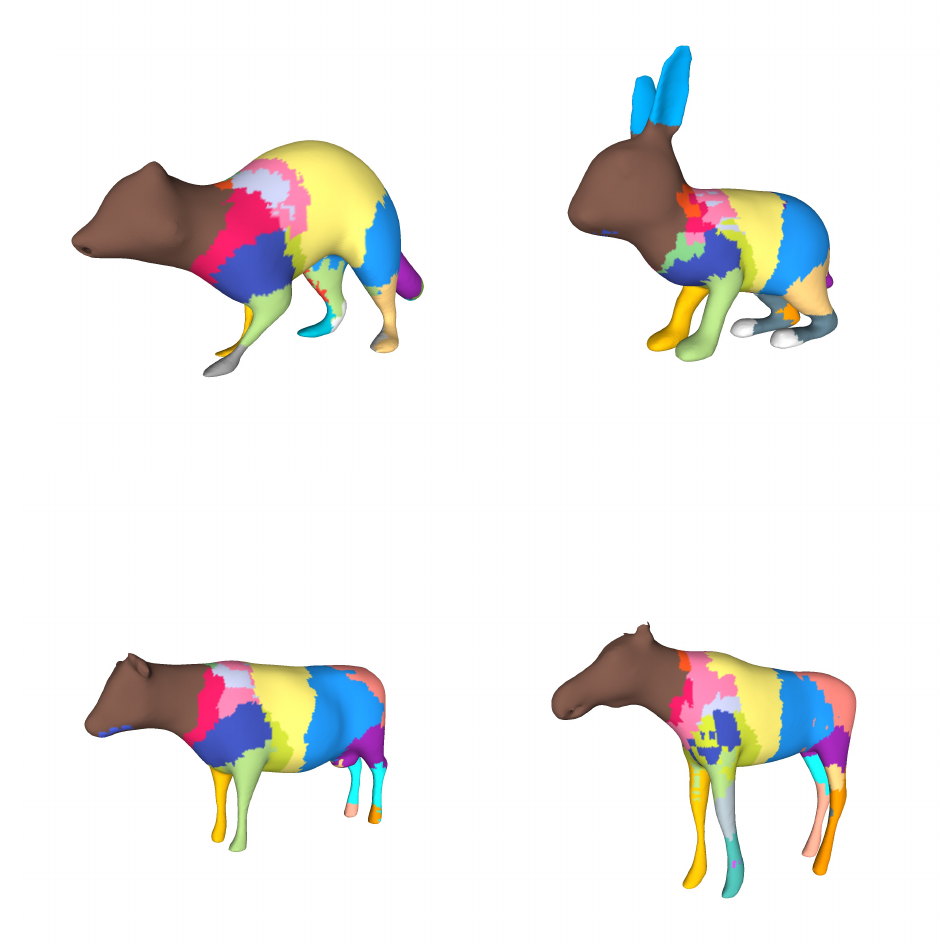}
  \caption{Skinning weights predicted by the skinning predictor module of our deformation representation.
  }
  \label{skinning_weights}
\end{figure}

\section{Skinning Weights}

To show how our deformation representation works, we visualized the skinning weights of mesh vertices predicted by the skinning predictor in \Cref{skinning_weights}. Colors are applied based on the class with the highest weight, and the shapes used for visualization are the source shapes employed in the motion retargeting experiments.

\begin{figure}[t]
  \centering
\includegraphics[width=0.8\linewidth]{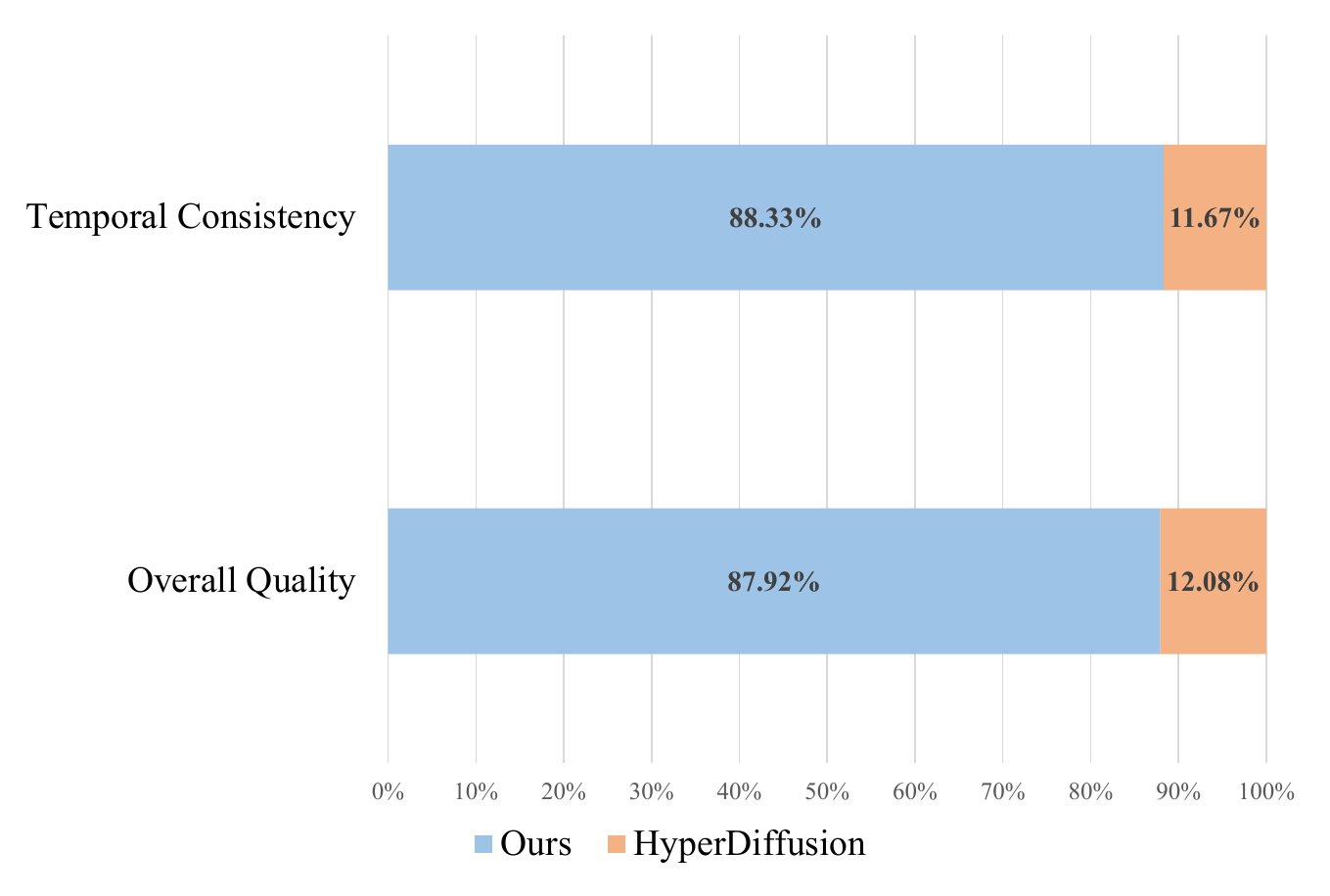}
  \caption{Human evaluation results for the qualitative evaluation of our model.
  }
  \label{human_evaluation}
\end{figure}

\section{Human Evaluation}

For the qualitative assessment of the generated 4D dynamic shapes, we conduct a human evaluation. We designed the experiment so that 30 evaluators each assess 8 pairs of videos for temporal consistency and overall quality. \cref{human_evaluation} shows that our method is preferred over HyperDiffusion \cite{Erkoc_2023_ICCV} by 88.33\% and 87.92\%, respectively, in terms of temporal consistency and overall quality.

\section{Additional Details}
In this section, We provide further details for the experiments included in the main manuscript. We additionally provide the code for more detailed information on the implementation of the model.

\noindent \textbf{Dataset preprocessing.} To get ground truth SDF values for query points, we normalize each shape of all frames into a size of $[-0.5, 0.5]^3$ and sample 200,000 points for the query points from the space. Additionally, we sample 100,000 surface points from the shapes and we can get two query points from each surface point by adding two noises sampled from two Gaussian distributions with mean of 0 and standard deviation of 0.01 and 0.001, respectively.

\noindent \textbf{Model architecture.} The latent dimension for the all modules of the SDF representation is 256, and the latent dimension for the all modules of 
the deformation representation is 64. The width of the tri-plane features of skinning predictor, shape encoder, and motion encoder is 128. The number of the layers of the SDF model and transformation decoder is 9, and the hidden dimension of the layers is 512. Transformations are represented by matrices in $\mathbb{R}^{k \times 7}$, which are formed by concatenating translations in $\mathbb{R}^{k \times 3}$ and quaternions representing rotations in $\mathbb{R}^{k \times 4}$. The transformation decoder predicts the transformations  from the concatenation of motion features in $\mathbb{R}^{k \times 64}$ and time matrices in $\mathbb{R}^{k \times 1}$ filled with the value $t$. 

\end{document}